# Image2Flow: A hybrid image and graph convolutional neural network for rapid patient-specific pulmonary artery segmentation and CFD flow field calculation from 3D cardiac MRI data

*Tina Yao[1], Endrit Pajaziti[1], Michael Quail[1], Silvia Schievano[1], Jennifer A Steeden[1][¶], Vivek Muthurangu[1][¶]\**

[1]Institute of Cardiovascular Science, University College London, London, UK

[¶]These authors contributed equally to this work.

*Corresponding Author

Email: v.muthurangu@ucl.ac.uk

## Abstract

Computational fluid dynamics (CFD) can be used for non-invasive evaluation of hemodynamics. However, its routine use is limited by labor-intensive manual segmentation, CFD mesh creation, and time-consuming simulation. This study aims to train a deep learning model to both generate patient-specific volume-meshes of the pulmonary artery from 3D cardiac MRI data and directly estimate CFD flow fields.

This study used 135 3D cardiac MRIs from both a public and private dataset. The pulmonary arteries in the MRIs were manually segmented and converted into volume-meshes. CFD simulations were performed on ground truth meshes and interpolated onto point-point correspondent meshes to create the ground truth dataset. The dataset was split 85/10/15 for training, validation and testing. Image2Flow, a hybrid image and graph convolutional neural network, was trained to transform a pulmonary artery template to patient-specific anatomy and CFD values. Image2Flow was evaluated in terms of segmentation and accuracy of CFD predicted was assessed using node-wise comparisons. Centerline comparisons of Image2Flow and CFD simulations performed using machine learning segmentation were also performed.

Image2Flow achieved excellent segmentation accuracy with a median Dice score of 0.9 (IQR: 0.86 – 0.92). The median node-wise normalized absolute error for pressure and velocity magnitude was 11.98% (IQR: 9.44–17.90%) and 8.06% (IQR: 7.54–10.41), respectively. Centerline analysis showed no significant difference between the Image2Flow and conventional CFD simulated on machine learning-generated volume-meshes.

This proof-of-concept study has shown it is possible to simultaneously perform patient specific volume-mesh based segmentation and pressure and flow field estimation using Image2Flow. Image2Flow completes segmentation and CFD in ~205ms, which ~7000 times faster than manual methods, making it more feasible in a clinical environment.

## Author summary

Computational fluid dynamics is an engineering tool that can be used in a clinical setting to non-invasively model blood flow through blood vessels, such as the pulmonary artery. This information can be used to inform treatment planning for patients, especially those with cardiovascular conditions. However, its routine use is limited by a labor-intensive process, requiring substantial expertise and computational resources. Recently, machine learning has offered solutions to automate parts of this process, but no single model has addressed the entire workflow.

Therefore, we created Image2Flow, a machine learning model capable of generating a patient-specific representation of a pulmonary artery while predicting blood flow through the vessel. Image2Flow can generate highly accurate pulmonary artery representations and reasonably accurate blood flow predictions. Notably, it performs this whole process in 205 ms, which is ~7000x faster than the manual methods. With its speed and accuracy, Image2Flow holds substantial promise for facilitating computational fluid dynamics in clinical settings.

# Introduction

Computational fluid dynamics (CFD) can aid the management of congenital heart disease (CHD) [1–3] through non-invasive estimation of hemodynamics (e.g. pressure gradients) and prediction of hemodynamic response to therapy. Several cardiovascular conditions have already been investigated using CFD, including coronary artery anomalies [4], aortic coarctation [5], tetralogy of Fallot [6], and univentricular hearts [7,8]. However, translation of CFD into the clinical environment is currently limited by: i) Time-consuming manual image segmentation [9], ii) CFD mesh generation, which is complex and often requires engineering expertise [10], and iii) The computationally intensive nature of CFD simulations, which results in long compute times [11,12].

Recently, deep learning (DL) approaches have been shown to provide accurate voxel-wise segmentation of computed tomography and magnetic resonance imaging (MRI) data [13]. Importantly, meshes derived from these DL segmentations provide comparable CFD results to human segmentations [14]. Nonetheless, voxel-wise segmentation can result in misclassified regions and other anatomical inconsistencies. Furthermore, mesh generation is still required, which requires some expertise and can be time consuming. Thus, graph-based neural networks that directly generate surface-meshes from 3D image data have also been investigated [15,16]. Notably, these mesh-based models outperform voxel-based methods in terms of both accuracy and the direct generation of smooth, CFD "simulation-suitable" meshes [17,18].

Although speeding up segmentation/mesh generation removes one of the barriers to clinical translation, CFD is still limited by long computation times. There have been studies that have used DL to speed up CFD. These include using anatomical shape descriptors as machine learning model inputs [19–22], as well as point-cloud [23,24] and graph-based [25] methods that act directly on meshes. However, there has been limited research into using DL to automate both the segmentation and CFD simulation process in a single model.

In this study, we propose to build on previous graph-based methods to simultaneously directly generate volume-meshes from 3D MRI and estimate pressure and velocity at each vertex. The aims of this study were to: i) Develop a DL model capable of taking a 3D cardiac MRI, creating a volume mesh reconstruction of the pulmonary artery, and predicting pointwise CFD-like pressure and velocity, ii) Evaluate segmentation accuracy by comparison

Evaluate the proposed DL model by comparing point-wise DL-predicted CFD pressure and flow with results obtained from a conventional CFD solver.

# Materials and methods

## Image2Flow model architecture

Our DL model — Image2Flow (Fig 1) — builds on the previously described MeshDeformNet, which combines an image convolutional encoding arm with a graph convolutional template transformation arm [16] to perform surface-based segmentation. Image2Flow improves on this by simultaneously performing volume-mesh segmentation and prediction of pressure and flow at each vertex.

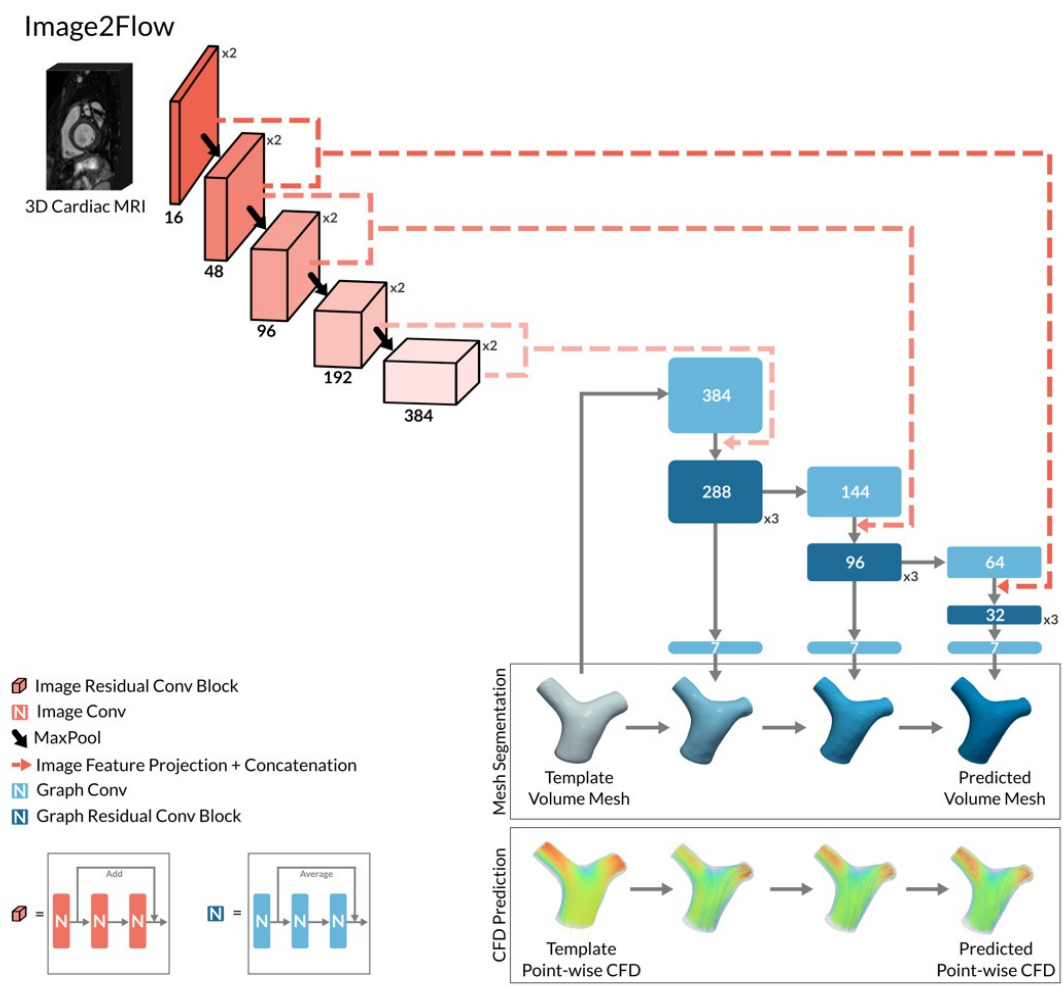

**Fig 1. The model architecture of Image2Flow.** *The hybrid image and graph convolutional neural network architecture of Image2Flow, it takes as input a 3D cardiac MRI and a template volume mesh of a pulmonary artery. It outputs the patient-specific pulmonary artery mesh with associated pressure and flow at each node.*

*Image encoding arm*

Image2Flow starts with an image encoding arm that takes a 3D MRI whole heart dataset as an input. The image encoding arm consists of five convolutional levels, each containing two convolutional residual blocks followed by a maxpool layer (Fig 1). Each convolutional residual block includes three sets of 3x3x3 convolution filter layers (followed by instance normalization and LeakyReLU activation), as well as a spatial dropout layer and a residual connection. The number of convolutional filters increases at each level (16, 48, 96, 192, and 384).

*Graph Transformation Arm*

The patient-specific output of Image2Flow was generated using the graph transformation arm, which utilizes features from the image encoding arm to transform a template mesh. The template volume-mesh was represented by a graph containing 10,998 nodes each with seven features (x-y-z coordinates, pressure, and flow), (Fig 1). The graph transformation arm comprised of three sequential transformation branches, each performing graph convolutions that successively convert the template into a patient-specific volume-mesh with estimated pressures and velocities at each vertex. We employed first-order Chebyshev convolutional layers from the Spektral library [26] for all graph convolutions.

Each graph transformation branch included the following steps: i) a graph convolution layer that increased/decreased the number of graph features to better match the number of image features in the corresponding levels of the image encoding arm, ii) projection of image features from two levels of the image encoding arm onto mesh nodes via trilinear interpolation, iii) concatenation of the extracted image features with graph features, iv) processing of concatenated features through a series of three graph residual convolutional blocks, and v) a bottleneck graph convolutional layer producing the transformation from the template mesh in terms of nodal x-y-z coordinates, pressure, and flow. Each graph convolution residual block consists of three graph convolutional layers (followed by instance normalization and LeakyReLU activation) and residual connection (Fig 1).

## Image data and processing

The training/validation/testing image dataset consisted of 135 cardiac triggered, respiratory navigated, 3D whole heart, balanced steady state free precession (WH-bSSFP) acquisitions. Of the 135 datasets, 83 datasets were collected from previously scanned children and adults with pediatric or CHD (excluding patients with single ventricles) as previously described [14]. This non-public dataset was ethically approved and obtained with written consent (Ref: 06/Q0508/124). The remaining 52 datasets were derived from the public multi-modality whole heart segmentation challenge (MMWHS) [27] data (n=60), which included a wide range of pathologies – 8 images excluded due to poor pulmonary artery definition.

All data were acquired with a 1.5T field strength. The private dataset exclusively comprised scans from Siemens scanners, while the public dataset included a combination of Siemens and Phillips scanned data. In the private dataset, scans were acquired isotropically with a voxel size of 1.6 mm. However, in the public dataset, voxel sizes were ~0.8-1.0 mm in-plane and ~1.0-1.6 mm through-plane. The total dataset was divided into 110/10/15 for training/validation/testing. The training set comprised a mix of non-public and public data, whereas the validation and test set exclusively used images randomly selected from the public dataset to enable public evaluation.

Reference standard conventional segmentation of the pulmonary arteries was performed by a single observer (20 years' experience in cardiac MRI post-processing) using a semi-automatic technique with manual correction (Plug-ins created in Horos v4.0, Horosproject.org, Maryland, USA). Initial segmentation was done using the fast level-set method [25]. This required the user to: i) set a threshold, ii) place seeds in the vessel of interest and iii) add blocking regions to prevent segmentation of unwanted structures. Manual correction of this initial segmentation was always required to remove unwanted structures, and this was performed using manual volume subtraction method in Horos. The WH-bSSFP data and their corresponding pulmonary artery masks were spline interpolated to create isotropic volumes with a voxel size of 1.0 mm. Both the interpolated image and mask data were centered around the position of the pulmonary artery and cropped to a 128x128x128 matrix.

### Point-point volume mesh generation

A prerequisite of Image2Flow was that all the patient-specific volume-meshes (and the template mesh) contained the same number of nodes with point-point correspondence. This enabled point-wise calculation of CFD losses during training, which was necessary to ensure accurate pressure and velocity estimation. This was achieved in a multi-step process shown in Fig 2 and described below.

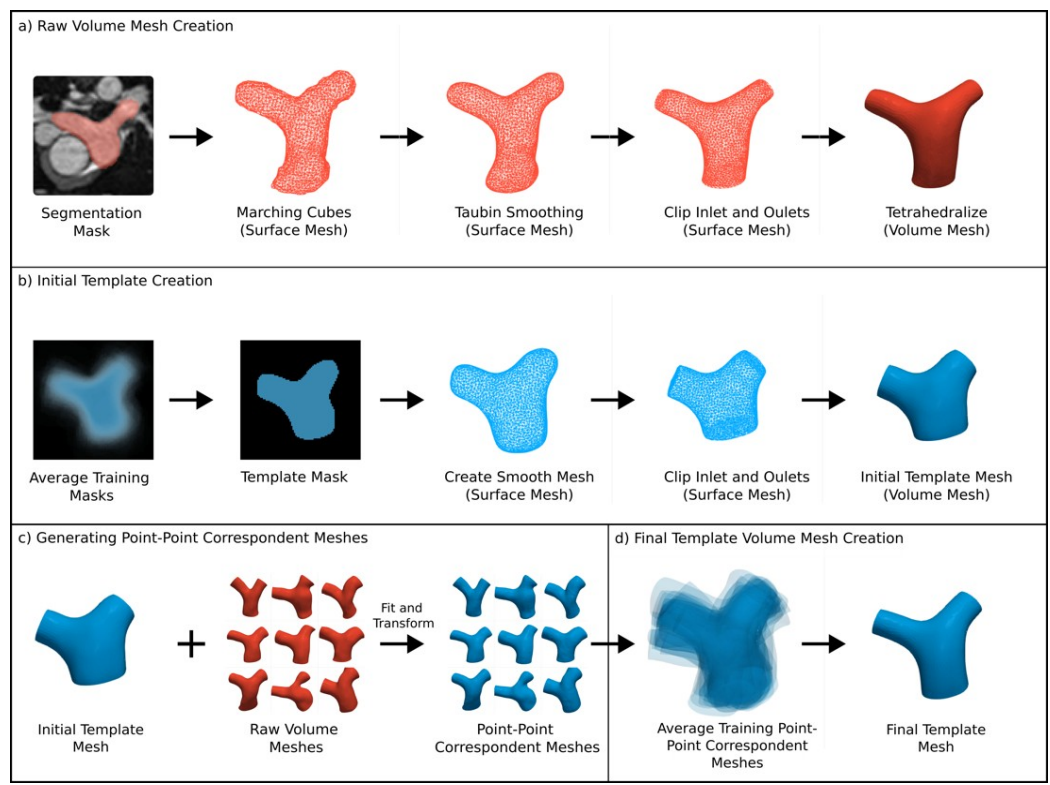

*Fig 2. Point-point Volume Mesh Generation.* A schematic of the steps involved in point-point correspondent volume mesh generation: (A) raw patient-specific volume mesh created from manual segmentation, (B) initial template volume mesh creation, (C) point-point correspondent volume mesh

*generation by transforming the initial template, (D) final template volume mesh creation by averaging the point-point correspondent meshes of the training data. Red indicates non-corresponding meshes and blue represents corresponding meshes. The wireframe rendering denotes surface meshes, while the solid rendering denotes volume meshes.*

In the first step, Fig 2A, raw patient-specific surface-meshes were generated from the segmentation masks using the marching cubes algorithm with subsequent smoothing. The raw surfaces were then manually clipped at the sino-tubular junction (inlet) and at the hilar branches of the pulmonary arteries (outlets). The clipped patient specific surface-meshes were defined as the ground truth (GT) surfaces. They were then converted into raw tetrahedral volume meshes using the TetGen in Vascular Modeling Toolkit (VMTK) [28]. To achieve approximate uniformity across meshes, the target edge length parameter for converting the surfaces into tetrahedral volume-meshes was adjusted iteratively until each mesh consisted of approximately ~11,000 (range: 10,890 – 11,922) nodes.

In the second step, Fig 2B, an initial pulmonary artery template was generated by averaging the segmentation masks from the training data and applying morphological image operations to generate a more consistent average template mask. The template mask was converted into a surface and clipped at inlet and outlets, from which a volume mesh was generated with 10,998 nodes using TetGen.

In the third step, Fig 2C, the initial template volume-mesh (generated in step 2) was deformed to match the patient-specific raw volume meshes. This resulted in patient specific volume-meshes that had the same number of nodes, all in point-point correspondence. The deformation process consisted of training and fitting a simplified version of our Image2Flow model (excluding blood pressure and flow features and CFD losses) to all datasets. The average symmetric surface distance (ASSD) and Hausdorff distance (HD) between the ground truth surfaces and the surfaces extracted from the point-point correspondent meshes were 0.148 (0.122 - 0.171) and 1.65 (1.465 - 1.873) respectively, The Dice score between the original masks and masks derived the point-point correspondent meshes was 0.976 (0.972–0.980). These results demonstrated the robustness of deformation process.

In the last step, Fig 2D, a final template mesh was created by averaging the node positions of the patient-specific point correspondent training meshes (generated in step 3). This final template mesh more accurately represented the mean shape of the pulmonary artery than the initial template mesh (from step 1) and was used for training the final model.

## Computational fluid dynamics (CFD) simulations

Neither the raw or point-point correspondent volume-meshes were suitable for CFD computation as finer meshes and flow extensions are required. Therefore, the GT surfaces were also used to create CFD-suitable volume meshes. Firstly, using VMTK, 40mm flow extensions were added to the clipped inlet and outlets of the surface, to create flat and circular cross-sections to achieve a uniform velocity profile. Subsequently, the extended surfaces were transformed into high-resolution volume meshes using TetGen, each containing an average of approximately 500,000 tetrahedral elements. The number of tetrahedral elements was chosen after sensitivity analysis (see S1 Fig).

CFD simulations were conducted across the refined meshes for all datasets (including the template) using Fluent (Ansys, Pennsylvania, USA). The simulations involved a steady inflow velocity of 0.2 m/s, which is the average velocity through the PA over the cardiac cycle based on the higher range of normal [29], while the gauge pressures at the outlets were set to 0 Pa. A uniform set of blood properties was applied, assuming blood to be a Newtonian fluid with a density of 1060 kg/m³ and a viscosity of 0.04 Pa·s. Additionally, laminar, steady-state flow conditions were employed, and standard non-slip conditions were maintained at the wall boundaries.

CFD simulations were performed on the meshes lacking point-point correspondence and with inconsistent numbers of points. Therefore, the CFD results required interpolation onto the point-correspondent meshes (including the template) for pressure/flow data alignment across cases, using the method described by Pajaziti et al. [19]. The accuracy of the pressure and flow values interpolation was 0.05% (-0.2 – 0.4) and -0.003% (-0.1 – 0.1) for pressure and velocity.

## Model training

Image2Flow was trained with a NVIDIA GeForce RTX 3090 GPU card (24 GB RAM). Due to the highly skewed pressure data within and between patients, we applied a cube root transformation to normalize the distribution. Pressure (after cube rooting) and flow were then scaled by subtracting the mean and dividing by the standard deviation. After inference, the predicted pressure was cubed again to preserve the original data scale. Our model was trained with several losses to optimize segmentation accuracy, volume-mesh quality and CFD pressure/flow prediction, as described below.

### Point loss

A Chamfer loss was used to minimize the point-wise distance between the template PA mesh and the ground truth. We applied two Chamfer losses: one for all points ($\mathcal{L}_{Point}$) within the mesh and another specifically for surface points ($\mathcal{L}_{Point,S}$):

$$L_{Point}(P,G) = \sum_{p \in P} \min_{g \in G} \|p-q\|_2^2 + \sum_{g \in G} \min_{p \in P} \|p-q\|_2^2$$

Where **p** and **g** are vertices in the point clouds (or surface point clouds) of the predicted and ground truth volume meshes **P** and **G**.

### Edge length deviation loss

We define the edge length deviation ($D_{Edge}$) as the mean edge length in the mesh relative to their standard deviation. This is a measure of the consistency of the edge lengths throughout the mesh. Since all the volume-meshes were generated via TetGen, the edge length deviation of the ground truth meshes are similar to the template's edge length deviation. Therefore, we calculate the edge length deviation loss by subtracting the template's edge length deviation from the deviation of the predicted mesh. This loss ensures that the points in the template mesh translate uniformly. We applied two edge length deviation losses: one for all edges within the mesh ($\mathcal{L}_{\text{Edge}}$) and another specifically for edges on the surface ($\mathcal{L}_{\text{Edge,S}}$), which ensures a smooth mesh surface:

$$D_{Edge}(M) = \frac{\sigma_{Edge}}{\mu_{Edge}}$$

$$L_{Edge}(P,T) = D_{Edge}(P) - D_{Edge}(T)$$

Where $\mu_{Edge}$ and $\sigma_{Edge}$ represents the mean and standard deviation of the edge lengths (or surface edge lengths) in volume mesh **M**. **T** represents the template volume mesh.

### Aspect ratio loss

Aspect ratio (*A*) is a key measure of volume mesh quality, representing the ratio of the longest to the shortest edge in each tetrahedral element. Since all volume meshes follow the same generation process, the aspect ratios of the ground truth meshes closely resemble the template's aspect ratio. Therefore, we calculate this loss by subtracting the template's aspect ratio from the predicted value. Incorporating the aspect ratio loss ($\mathcal{L}_{\text{Aspect}}$) guarantees low-skew, consistent tetrahedral elements in predicted volume meshes and prevents self-intersecting faces:

$$A(M) = \frac{1}{|C|} \sum_{C_j \in C} \frac{\max_{\forall v_j \neq v_k \in c_j} \|v_j - v_k\|_2}{\min_{\forall v_j \neq v_k \in c_j} \|v_j - v_k\|_2}$$

$$L_{Aspect}(P,T) = A(P) - A(T)$$

Where $C_j$ represents the *j*th tetrahedral cell element in the set of tetrahedral elements, C, within the volume mesh **M**. The set of vertices in $C_j$ is denoted $c_j$, and **v**$_j$, **v**$_k$ represent vertices in $c_j$.

### Cap coplanar loss

Kong et al. outlined the "cap coplanar loss", which ensures flat surfaces at the inlets and outlets of a mesh, which is desirable for CFD simulation. [17] This loss ($\mathcal{L}_{\text{Cap}}$) minimized the normals to the faces on the cap, aligning the surface faces in the same direction.

$$L_{Cap}(P) = \sum_{j=1}^{3} \sum_{k=F_j} \left\| n_k - \frac{1}{|F_j|} \sum_{k=F_j} n_k \right\|_2^2$$

Where $F_j$ is the mesh faces of the *j*th cap in the predicted volume mesh **P**. There are three caps, representing the inlet and the two outlets. ***n***$_k$ is the normal vector of the *j*th face of F$_j$.

### CFD loss

By registering the ground truth meshes with the template meshes, we established point correspondence across all meshes. Consequently, when we interpolated CFD results onto these meshes, the blood pressures and flows also aligned. This alignment enabled the calculation of the mean absolute error loss, between ground truth and predicted pressure and velocity values ($\mathcal{L}_{CFD}$). Standardization of pressure and flow values ensured they followed a similar scale, enabling their combination into a single CFD loss through summation.

$$L_{CFD}(P,G) = \frac{1}{N} \sum_{j=1}^{N} |P_j - G_j|$$

Where $P_j$ and $G_j$ represent the pressure and flow value at the vertex $j$ of the volume meshes **P** and **G**.

### Total loss

At each of the three transformation branches, the model generated a mesh with associated pointwise CFD pressure and flow. This allowed calculation of mesh loss ($\mathcal{L}_{mesh}$) for each transformation branch by combination of the losses:

$$L_{mesh} = \lambda_1(L_{point} + L_{point,S}) + \lambda_2(L_{Edge} + L_{Edge,S}) + \lambda_3 L_{Aspect} + \lambda_4 L_{Cap} + \lambda_5 L_{CFD}$$

The individual losses were weighted empirically ($\lambda_1 = 1$, $\lambda_2 = 0.5$, $\lambda_3 = 1.25$, $\lambda_4 = 0.005$ and $\lambda_5 = 15$) to create a high mesh quality, smooth, and accurate segmentation with flat inlet and outlets while also achieving the highest pressure/flow accuracy.

The total loss is the sum of losses of the three transformation branches:

$$L_{Total} = \sum_{i=1}^{3} L_{Mesh}(P_i, G, T)$$

Where **P, G** and **T** are the predicted, ground truth and template volume meshes, respectively.

### Inference

At inference, 3D WH-bSSFP images from 15 unseen cases were inputted into the Image2Flow. The outputs were volume-meshes with each vertex associated with pressure and velocity values. In addition, to better understand the origin of any errors, the output volume mesh segmentations from Image2Flow (without pressure and velocity values) were also used to perform standard CFD as previously described (CFD$_{DL-seg}$). Inference time was measured as the time for prediction on a NVIDIA GeForce RTX 3090 GPU card (24GB RAM).

## Evaluation

### Segmentation accuracy

Segmentation accuracy was assessed using Dice score, ASSD, and Hausdorff distance (HD). For benchmarking, we compared Image2Flow with our implementation of the MeshDeformNet [16], and a 3D UNet model as described by Montalt-Tordera et al. [14]. Our 'MeshDeformNet' is essentially the surface-mesh version of Image2Flow, employing the same losses except for the aspect ratio loss, which is specific to volume-meshes. The surface-meshes that were created had the same number of nodes as the volume-meshes (10,998).

As Dice score compares segmentation masks, we first converted the meshes predicted by Image2Flow and 'MeshDeformNet' into binary masks (the 3D UNet outputs were already binary mask). The Dice score was then calculated by comparison with the ground truth manual segmentation masks. As ASSD and HD are both computed on surfaces, the segmentation masks from the 3D UNet were first converted into surfaces and the surfaces were extracted from the Image2Flow volume-meshes (the 'MeshDeformNet' outputs were already surfaces). The ASSD and HD were then calculated by comparison with the ground truth surfaces.

### *CFD accuracy*

Node-wise comparisons between $CFD_{I2F}$ and the non-interpolated ground truth CFD data ($CFD_{GT}$) were not possible due to the lack of point-point correspondence and the different number of mesh nodes. Thus, CFD accuracy was evaluated in two ways:

Node-wise error: The ground truth CFD results were interpolated onto the point-point correspondent mesh ($CFD_{GT-INT}$). At each node, we then calculated the normalized absolute error (NAE) for pressure, x-y-z velocity components and velocity magnitude as previously described [19]. Mean normalized absolute error ($MNAE_s$), the average NAE value for each subject, was calculated to determine the error over the whole mesh. Conversely, mean node error ($MNAE_n$), which is the average NAE value across the population of each node, was used to evaluate the prediction error with respect to their relative position in the mesh.

Centerline error: Additionally, we used the Fréchet Distance (FD), which measures the similarity between two curves [30] has previously been used to evaluate the accuracy of DL predicted CFD pressure/flow gradients distributions [19]. In this study, we used FD to compare pressure and velocity magnitude values along the $CFD_{I2F}$ and $CFD_{GT}$ centerlines.

For the centerline analysis, we computed the centerlines for the ground truth and the predicted meshes using VMTK for both the left and right pulmonary arteries. The centerlines were resampled to 100 equally spaced points to enable comparison of the predicted and ground truth CFD values. Pressure and velocity magnitude values along the centerline were obtained by averaging the values of the five closest nodes in the mesh to each resampled centerline point. We were also interested in investigating Image2Flow's CFD prediction capability independently of segmentation accuracy. Therefore, we also compared $CFD_{DL-seg}$ to $CFD_{GT}$ using FD after the same centerline analysis.

As FD values do not adequately account for the wide range of pressure and velocity values, we normalized FD ($FD_{norm}$) by dividing the FD by the range of pressure/velocity in $CFD_{GT}$ on a per-patient basis. $FD_{norm}$ was computed between $CFD_{I2F}$ and $CFD_{GT}$, as well as between $CFD_{I2F-seg}$ and $CFD_{GT}$.

*Statistics*

Continuous variables are presented as median (interquartile range). Bland-Altman analysis evaluated bias and limits of agreement for pressure and flow predictions. The Wilcoxon signed-rank test compared segmentation performance between Image2Flow, 'MeshDeformNet' and the 3D UNet. The centerline analysis also used the Wilcoxon signed-rank test to compare $FD_{norm}$ for $CFD_{I2F}$ and $CFD_{DL-seg}$. The Wilcoxon signed-rank test was used as the differences between the measurements were not normally distributed (evaluated using the Shapiro-Wilk test). P-values less than .05 were considered statistically significant.

# Results

## Feasibility

Image2Flow was able to successfully patient-specific volume-meshes with pressure and flow estimations for all subjects. The inference time was approximately 205ms per dataset. This represents a speed-up of ~7000x, with conventional segmentation, processing and simulation taking ~25 minutes in total (segmentation: 15min, mesh generation: 5min, CFD simulation: 5min).

## Segmentation Accuracy

Fig 3 visualizes the best, median and worst Image2Flow segmentations (in terms of Dice score) and compares with the 'MeshDeformNet' like model and a 3D UNet. Compared to the 3D UNet, Image2Flow produces smooth, anatomically correct meshes with well-defined flat inlets and outlets and without misclassified islands. This is also corroborated quantitatively (Table 1), with Image2Flow segmentations having significantly better HD scores than the 3D UNet (6.71 [5.36– 7.46] vs. 9.87 [7.60 – 15.70], p=.0004). It should be noted that 'MeshDeformNet' also produced smooth meshes that outperformed the 3D UNet in terms of HD score (p=.0002). However, Image2Flow had a significantly higher Dice score compared to the 'MeshDeformNet' (0.90 [0.86 – 0.92] vs. 0.89 [0.84 – 0.90], p =.005).

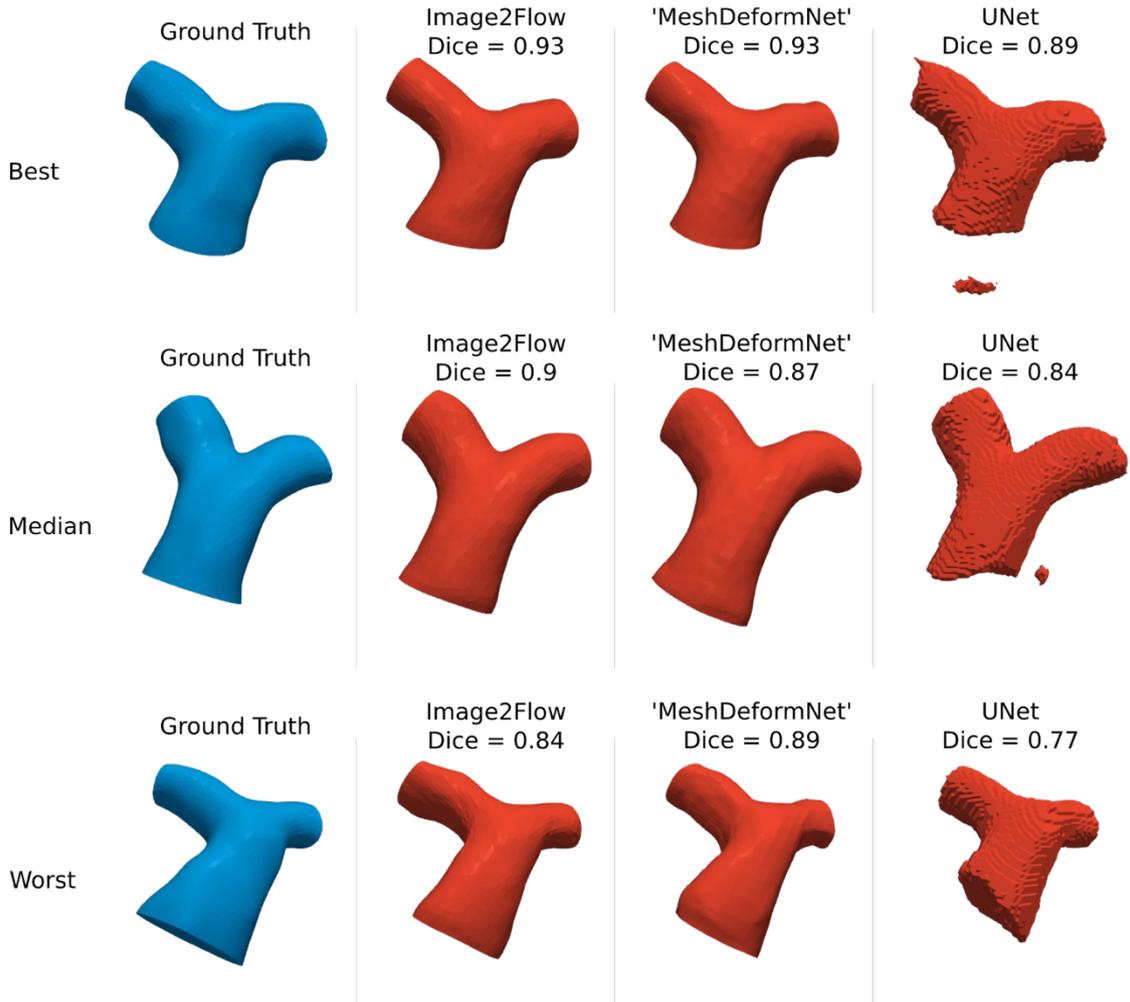

*Fig 3. Segmentation accuracy.* The best, median and worst volume mesh segmentation of Image2Flow compared to 'MeshDeformNet' and a 3D UNet.

*Table 1. Segmentation metrics evaluating Image2Flow, 'MeshDeformNet' and the 3D UNet compared to the ground truth.*

| Metric | Image2Flow | 'MeshDeformNet' | 3D UNet |
|---|---|---|---|
| **Dice (↑)** | 0.90 (0.86 – 0.92) ** | 0.88 (0.84 – 0.90) | 0.89 (0.84 – 0.90) |
| **ASSD (mm) (↓)** | 1.46 (1.25 – 1.59) | 1.38 (1.12 – 1.73) | 1.47 (1.34 – 1.52) |
| **HD (mm) (↓)** | 6.71 (5.36– 7.46) * | 6.99 (5.35 – 7.79) *** | 9.87 (7.60 – 15.70) |

\* Indicates statistical significance comparing Image2Flow – UNet. ** indicates statistical significance comparing Image2Flow – 'MeshDeformNet'. *** indicates statistical significance comparing 'MeshDeformNet' – UNet

## CFD Comparison: Node-wise

Subject-level $MNAE_s$ ($CFD_{I2F}$ vs. $CFD_{GT\text{-}INT}$) for pressure and velocity are shown in Table 2, with errors of ~12% for pressure and between ~4-8% for velocity components. In Fig 4, the best, median, and worst pressure and velocity predictions are shown. Even in the worst case, Image2Flow conserves pressure distributions and velocity streamline patterns.

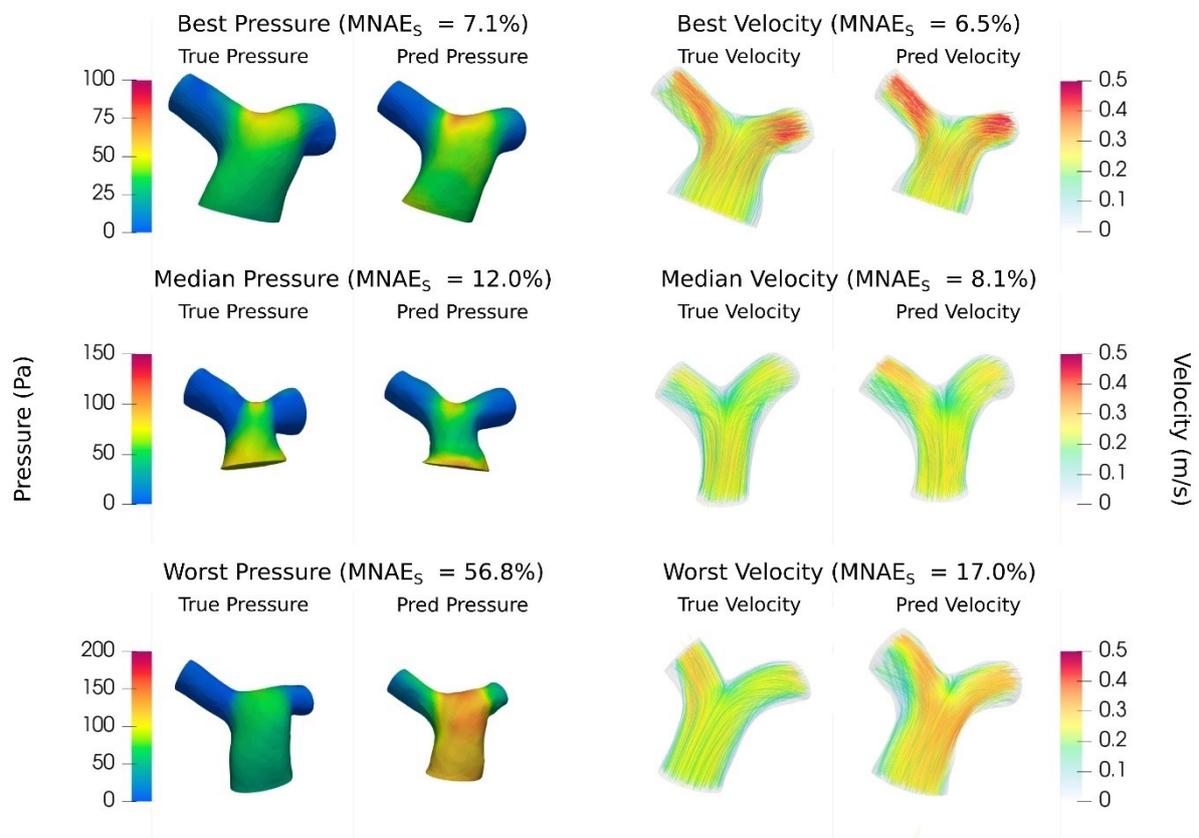

**Fig 4. Subject-level CFD prediction error.** *The best, median and worst blood pressure and flow predictions of Image2Flow by $MNAE_s$. The size and positioning between the true and predicted meshes are to scale.*

**Table 2. Subject-level CFD prediction error for Image2Flow compared to the ground truth CFD interpolated onto point-point correspondent meshes.**

| CFD Variable | $MNAE_s$ (%) | RMSE |
| --- | --- | --- |
| **Pressure (Pa)** | 11.98 (9.44 – 17.90) | 11.42 (7.72 – 36.16) |
| **X-Velocity (m/s)** | 6.34 (5.00 – 7.21) | 0.027 (0.024 – 0.036) |
| **Y-Velocity (m/s)** | 7.14 (5.84 – 8.69) | 0.038 (0.029 – 0.053) |
| **Z-Velocity (m/s)** | 4.17 (3.99 – 4.66) | 0.033 (0.030 – 0.046) |
| **Velocity-Magnitude (m/s)** | 8.06 (7.54 – 10.41) | 0.046 (0.036 – 0.062) |

Fig 5A visualizes the distribution of node errors across the whole population for pressure and velocity. Errors are not higher in any specific location (e.g. the bifurcation), demonstrating that Image2Flow CFD predictions have no spatially localized biases. Across the whole pulmonary artery, Bland-Altman plots (Fig 5B) do show a small but significant bias (p<.0001) in pressure (0.27%) and velocity predictions (-1.53%). Similarly, there are minimal but significant (p<.0001) biases in the x-y-z velocity components of 0.04%, -1.30% and -0.81%, respectively (see S2 Fig).

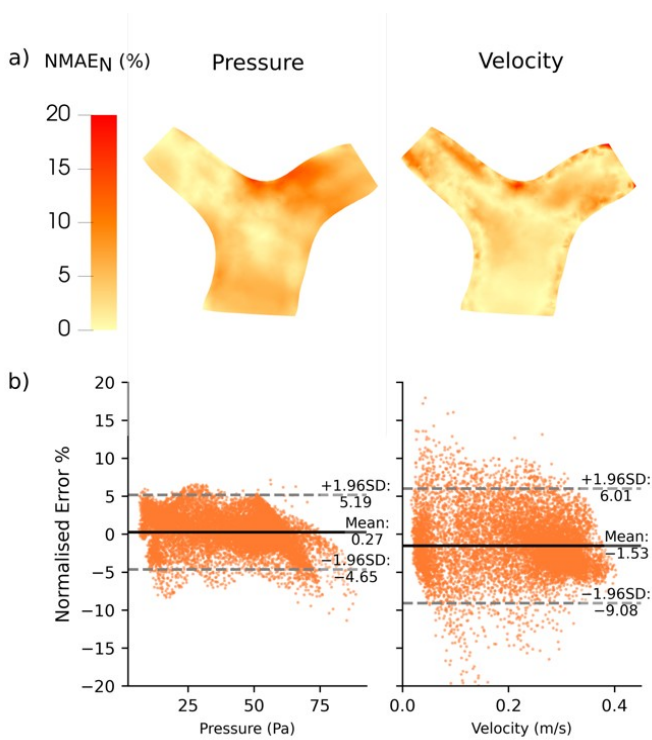

**Fig 5. CFD prediction error distribution across the pulmonary artery.** *The distribution of node-wise error ($MNAE_N$) of the test set (n = 15) projected onto the template pulmonary artery volume-mesh. (A) Distribution of error across the cross-section of the pulmonary artery, (B) Bland-Altman analysis of the pressure and velocity magnitude errors*

## CFD Comparison: Centerline

Fig 6 and 7 shows pressure and velocity curves along the LPA and RPA centerlines for the best, median and worst cases by $FD_{norm}$. For velocity, although $CFD_{DL-seg}$ was slightly closer to the ground truth than $CFD_{I2F}$, these differences did not reach statistical significance (p>.12).

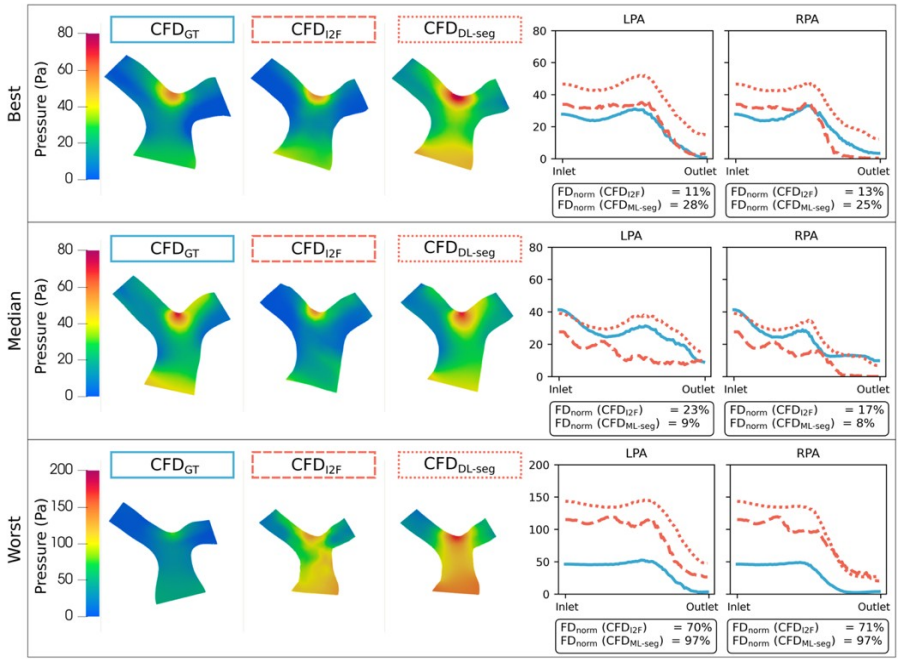

***Fig 6. Pressure centerline error analysis.*** *Centerline error analysis of pressure gradients in the left and right pulmonary arteries. The best, median and worst normalised Frechet Distance of Image2Flow are shown compared with the ground truth CFD and conventional CFD simulated on the volume-mesh generated from deep learning segmentation.*

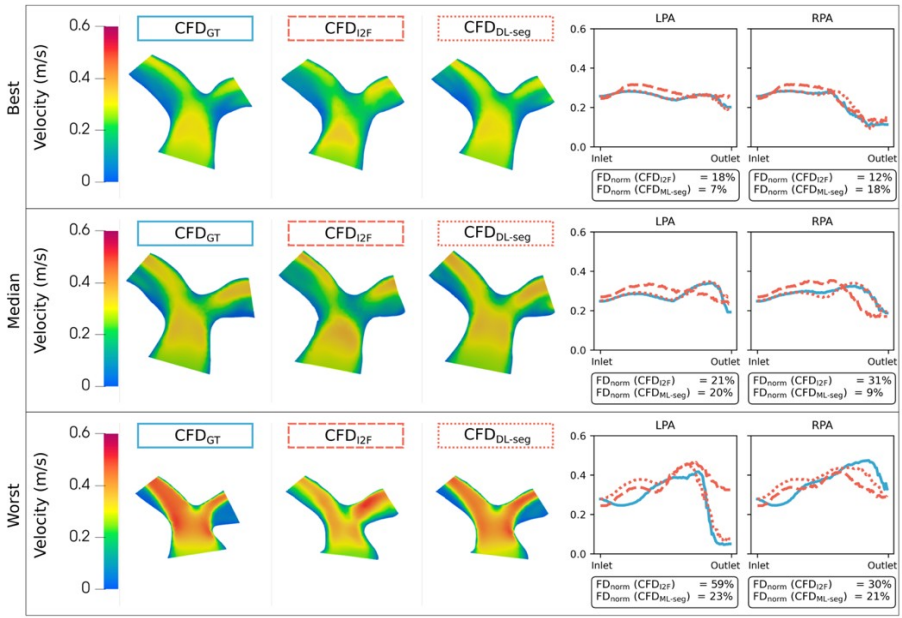

***Fig 7. Velocity centerline error analysis*** *Centerline error analysis of velocity magnitude gradients in the left and right pulmonary arteries. The best, median and worst normalised Frechet Distance of Image2Flow are shown compared with the ground truth CFD and conventional CFD simulated on the volume-mesh generated from deep learning segmentation.*

For pressure, the comparison between $CFD_{I2F}$ and $CFD_{DL-seg}$ was more complex. In the best case, $CFD_{I2F}$ outperformed $CFD_{DL-seg}$, but the opposite was true in the median case, and both $CFD_{I2F}$ and $CFD_{DL-seg}$ performed poorly in the worst case. As a result, there were no statistically significant differences in $FD_{norm}$ (Table 3) between $CFD_{I2F}$ and $CFD_{DL-seg}$ for either

LPA or RPA (p>.42). It should also be noted that there were also no statistical differences in $FD_{norm}$ (for pressure and velocity) between the LPA and RPA for either $CFD_{I2F}$ (p>0.33) or $CFD_{DL-seg}$ (p>.45).

Table 3. Normalized Frechet distances (%) for centerline pressure and velocity between Image2Flow predicted CFD and conventional CFD simulated on deep learning volume-mesh segmentation versus ground truth CFD.

|  | $CFD_{GT}$ vs $CFD_{I2F}$ | | $CFD_{GT}$ vs $CFD_{DL-seg}$ | |
| --- | --- | --- | --- | --- |
|  | LPA | RPA | LPA | RPA |
| **Pressure** | 20.8 (13.3–14.9) | 20.2 (15.4–26.8) | 20.5 (12.2–23.6) | 19.4 (10.8–23.6) |
| **Velocity** | 20.4 (16.4–28.8) | 27.6 (23.9–32.7) | 15.4 (11.1–22.7) | 24.9 (17.0–30.2) |

# Discussion

To our knowledge, this is the first study that uses a graph-based DL model to efficiently segment and estimate flow fields in the pulmonary arteries directly from 3D whole heart data. The main findings of this study were: i) Image2Flow can generate highly accurate pulmonary artery volume-meshes directly from cardiac MRI data, and ii) Image2Flow can also robustly estimate pressure and velocity at mesh vertices. The main benefit of Image2Flow is automatic and rapid inference (~205ms), which is approximately 7,000x faster than manual processing and conventional CFD simulation. We believe the ability to perform rapid 'CFD-like' evaluation could aid with clinical translation of hemodynamic simulation in structural and congenital heart disease.

## Segmentation Accuracy

Image2Flow builds on previous work that has shown that graph convolutional neural networks (e.g. MeshDeformNet) can generate surface-meshes with excellent segmentation accuracy. However, unlike previous surface-mesh based approaches, Image2Flow also estimates pressure and flow, and this requires the generation of a volume-mesh. To achieve this, we employed a modified network architecture and a volumetric template. We also utilized several novel losses that ensured the final volume-mesh had regularly shaped and sized cells (edge length deviation loss and aspect ratio loss), flat inlet and outlets (cap co-planar loss), and a smooth surface (surface edge length deviation loss). This resulted in a mesh that required no further processing (e.g. clipping). As with MeshDeformNet, Image2Flow performed better than a traditional 3D UNet, producing more anatomically consistent segmentations with flat inlet and outlets. Furthermore, Image2Flow slightly outperformed our 'MeshDeformNet' implementation in terms of Dice score, with no statistically significant differences in HD or ASSD. This demonstrates that generating volume meshes (rather than surface-meshes) does not have a deleterious impact on segmentation accuracy and may even improve robustness.

## CFD Comparison

Image2Flow was able to robustly estimate pressure and flow at each mesh vertex, with no spatially localized systematic biases in pressure or flow. However, compared to previous DL-CFD methods, Image2Flow did have higher node-wise errors. Pajaziti et al. presented a DL-CFD model for the aorta that achieved MNAEs of ~5% for pressure and ~4% for velocity, while Li et al. achieved MNAEs of ~2-3% for both pressure and velocity in the coronary arteries [19,23]. One factor contributing to lower errors in previous models is that they estimate pressure and velocity from pre-derived anatomy (shape descriptors or point clouds). On the other hand, Image2Flow attempts simultaneous segmentation and flow field estimation, and this is a more complex task. Another factor is the smaller amount of training data used in our study compared to previous studies. This was primarily because we were unable to easily produce synthetic data for training. Pajaziti et al. used 3000 synthetic datasets generated from a shape model in their study, which is 20 times larger than the data used in this paper [19]. Unfortunately, generating such datasets is challenging in our approach because synthetic images, rather than meshes, must be produced. Recently, generative adversarial models have been shown to effectively produce paired synthetic image and mesh-segmentation data [31]. We believe that such an approach could be used to generate larger training datasets and potentially improve the accuracy of Image2Flow.

Normalized Frechet distances along the vessel centerlines for pressure and velocity were used to gauge the accuracy of Image2Flow compared to the ground truth conventional CFD performed using manually segmented data. This analysis also enabled Image2Flow to be compared to conventional CFD performed using the Image2Flow segmentation. Interestingly, there was no statistical difference in normalized Frechet distances between Image2Flow output and Image2Flow segmentation with conventional CFD. Moreover, in some cases, the Image2Flow segmentation with conventional CFD had greater pressure errors than the Image2Flow output. This suggests that the model may learn to correct errors in anatomy when estimating the final pressure. It also suggests that small changes in the volume mesh may lead to greater variation in the pressure compared to the velocity, as noted by Montalt-Tordera et al. [14]. Expanding the training dataset and incorporating a broader range of anatomical variations could potentially improve the model's capacity to learn more complex relationships between geometry and hemodynamics. Nevertheless, Image2Flow does provide relatively accurate measures of hemodynamics in a fraction of the time of conventional means.

## Potential clinical applications

Image2Flow provides rapid segmentation and efficient hemodynamic computation, and we believe that this has significant potential in the clinical environment. We envisage a 'one-button CFD' approach performed by clinicians with no computational or engineering background as part of regular reporting of cardiovascular MRI studies. Routine, non-invasive estimation of hemodynamics should result in better identification of patients requiring intervention and, thus, potentially improved outcomes. However, this is a proof-of-concept study, and further improvements are required prior to any clinical validation (particularly the inclusion of patient-specific boundary conditions and time-varying flow fields—see limitations).

## Limitations

The main limitation of this study is that our model is trained only to predict a simplistic CFD where we used identical boundary conditions, velocity profiles and steady-state conditions for all patients. Morbiducci et al. [32] demonstrated that idealized boundary conditions can yield misleading representations of hemodynamics, emphasizing the necessity for patient-specific parameters to simulate realistic blood flow patterns in the complex cardiovascular system [33]. To address this, future studies should incorporate patient-specific flow profiles derived from phase-contrast MRI images, transition from steady-state to transient simulations, and use lumped parameter models at the outlets. Incorporating such information will require the development of novel graph architectures that both allow the prediction of dynamic hemodynamics and the incorporation of richer patient data.

Image2Flow also faces constraints on the number of nodes it can predict within the volume-meshes, which is currently limited to around 11,000 nodes due to GPU memory constraints (as also highlighted by Kong et al. [16]). Given the high-resolution volume meshes necessary for accurate blood flow calculations in CFD simulations, our node-wise evaluation is restricted to comparing low-resolution CFD predictions with ground truth results interpolated onto low-resolution volume meshes. Nonetheless, centerline analysis shows good agreement, suggesting that the current node limitation may be sufficient.

## Conclusion

Our proof-of-concept study introduces Image2Flow, a hybrid image and graph deep learning model for efficient pulmonary artery segmentation and flow field estimation from 3D whole heart data. Evaluated on 15 test cases, Image2Flow demonstrated superior segmentation accuracy compared to 3D UNet and MeshDeformNet, directly reconstructing anatomically correct volume meshes. The model exhibited robust pressure and velocity estimation shown through node-wise and centerline analysis.

Image2Flow excels in efficiency, producing flow fields over 7000x faster than conventional methods in a single pass, without requiring clinical and engineering expertise. This speed makes it promising for integration into a clinical setting, allowing swift patient evaluation for intervention and improved treatment outcome predictions. However, limitations and challenges remain, including the accuracy of point-point correspondence between nodes and the need for patient-specific parameters for more realistic CFD simulations.

# Supporting information

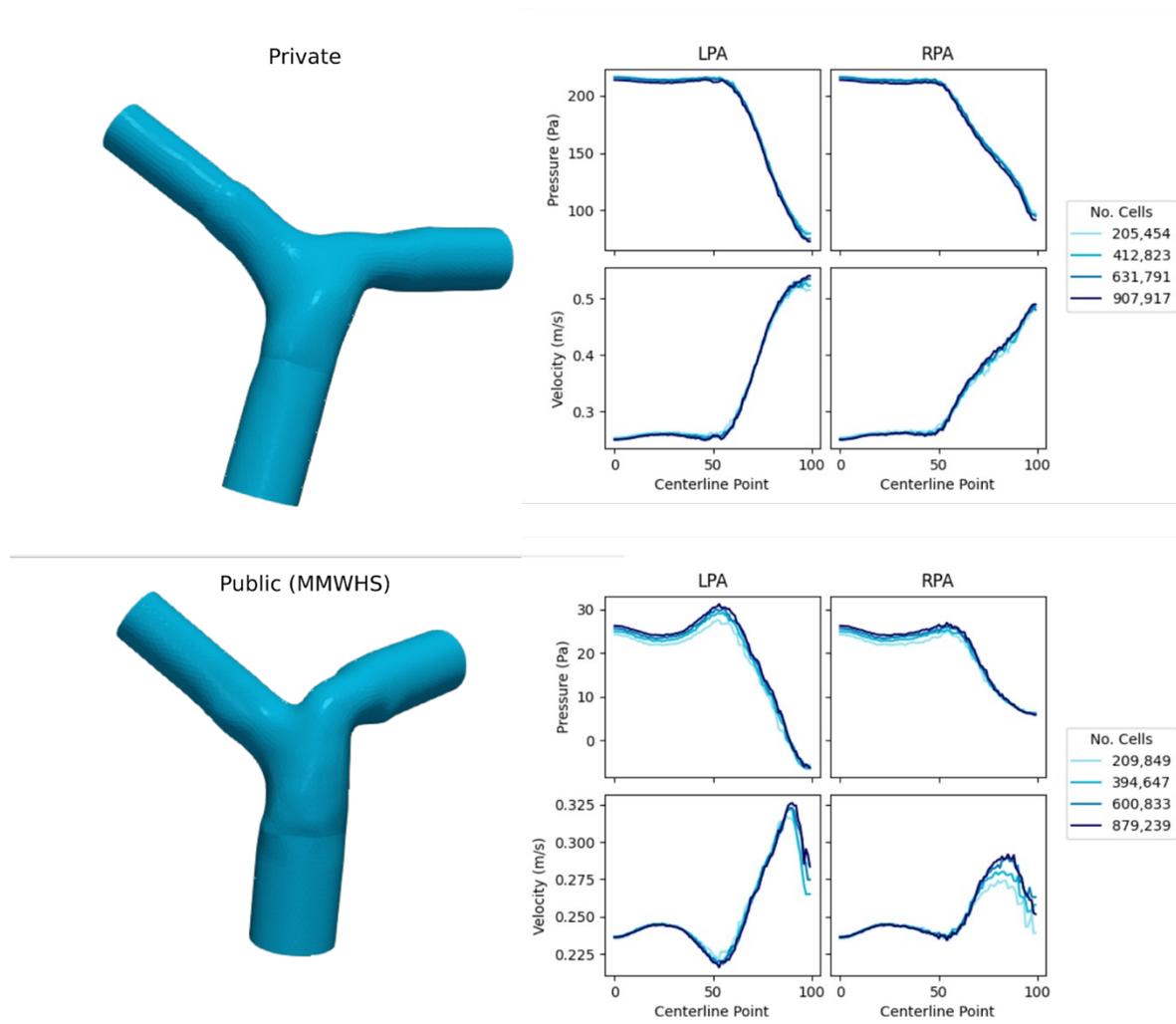

***S1 Fig. Sensitivity analysis.*** *Mesh sensitivity study conducted using two random pulmonary artery shapes sourced from private and public datasets. Volume meshes of approximately 200,000, 400,000, 600,000, and 800,000 cells were compared for each shape under identical CFD simulation boundary conditions, focusing on centerline pressure and velocity values. The analysis concluded with a decision to use 500,000 cells in the mesh to maintain a balance between accuracy, computation time, and memory usage.*

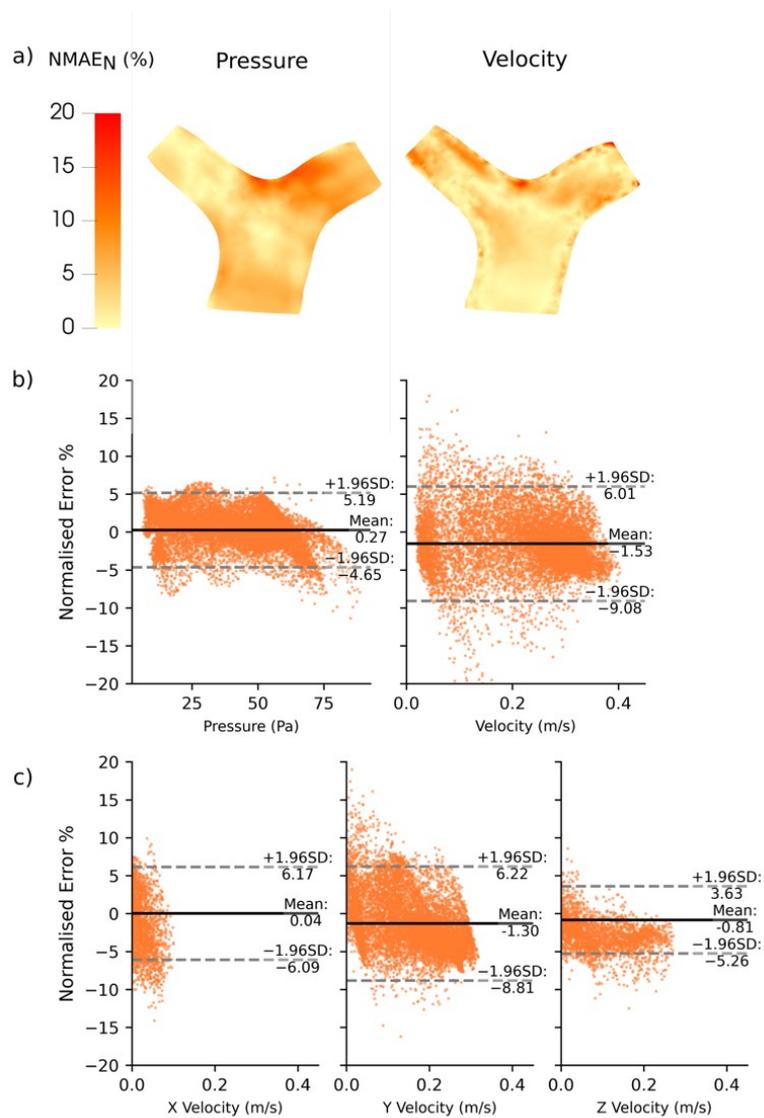

**S2 Fig. CFD prediction error distribution across the pulmonary artery.** The distribution of node-wise error (MNAEN) of the test set (n = 15) projected onto the template pulmonary artery volume-mesh. (A) Distribution of error across the cross-section of the pulmonary artery, (B) Bland-Altman analysis of the pressure and velocity magnitude errors, (C) Bland-Altman analysis of the errors of each of the x-y-z components of velocity